\documentclass{article}

\usepackage{PRIMEarxiv}

\usepackage[utf8]{inputenc} 
\usepackage[T1]{fontenc}    
\usepackage{hyperref}       
\usepackage{url}            
\usepackage{booktabs}       
\usepackage{amsfonts}       
\usepackage{nicefrac}       
\usepackage{microtype}      
\usepackage{lipsum}
\usepackage{fancyhdr}       
\usepackage{graphicx}       

\usepackage[linesnumbered, ruled, vlined]{algorithm2e}
\SetKwInput{KwInput}{Input}
\SetKwInput{KwOutput}{Output}

\usepackage{multirow}
\usepackage{amsmath}
\usepackage{subcaption}
\usepackage{float}
\usepackage[labelfont=bf]{caption}
\usepackage[table]{xcolor}
\usepackage{placeins}

\graphicspath{{media/}}     

\title{SemRAG: Semantic Knowledge-Augmented RAG for Improved Question-Answering
\thanks{\textit{\underline{Citation}}: 
\textbf{Kezhen Zhong. SEMRAG. Pages.... DOI:000000/11111.}} 
}

\author{
  Kezhen Zhong \\
  University of Sydney \\
  Sydney\\
  \texttt{kzho3733@uni.sydney.edu.au}
   \And
  Basem Suleiman, Shijing Chen \\
  University of New South Wales \\
  Sydney\\
  \texttt{b.suleiman@unsw.edu.au, arthur.chen@unsw.edu.au} \\
  \AND
  Abdelkarim Erradi \\
  Qatar University  \\
  Qatar \\
  \texttt{erradi@qu.edu.qa} \\
}

\begin{document}
\maketitle

\begin{abstract}
This paper introduces SemRAG, an enhanced Retrieval Augmented Generation (RAG) framework that efficiently integrates domain-specific knowledge using semantic chunking and knowledge graphs without extensive fine-tuning. Integrating domain-specific knowledge into large language models (LLMs) is crucial for improving their performance in specialized tasks. Yet, existing adaptations are computationally expensive, prone to overfitting and limit scalability. To address these challenges, SemRAG employs a semantic chunking algorithm that segments documents based on the cosine similarity from sentence embeddings, preserving semantic coherence while reducing computational overhead. Additionally, by structuring retrieved information into knowledge graphs, SemRAG captures relationships between entities, improving retrieval accuracy and contextual understanding. Experimental results on MultiHop RAG and Wikipedia datasets demonstrate SemRAG has significantly enhances the relevance and correctness of retrieved information from the Knowledge Graph, outperforming traditional RAG methods. Furthermore, we investigate the optimization of buffer sizes for different data corpus, as optimizing buffer sizes tailored to specific datasets can further improve retrieval performance, as integration of knowledge graphs strengthens entity relationships for better contextual comprehension. The primary advantage of SemRAG is its ability to create an efficient, accurate domain-specific LLM pipeline while avoiding resource-intensive fine-tuning. This makes it a practical and scalable approach aligned with sustainability goals, offering a viable solution for AI applications in domain-specific fields. By leveraging semantic chunking and knowledge graphs, SemRAG effectively enhances domain-specific LLM performance while minimizing computational demands, addressing a critical challenge in deploying advanced large language models efficiently.

\end{abstract}

\begin{figure}[ht]
  \includegraphics[width=0.7\linewidth]{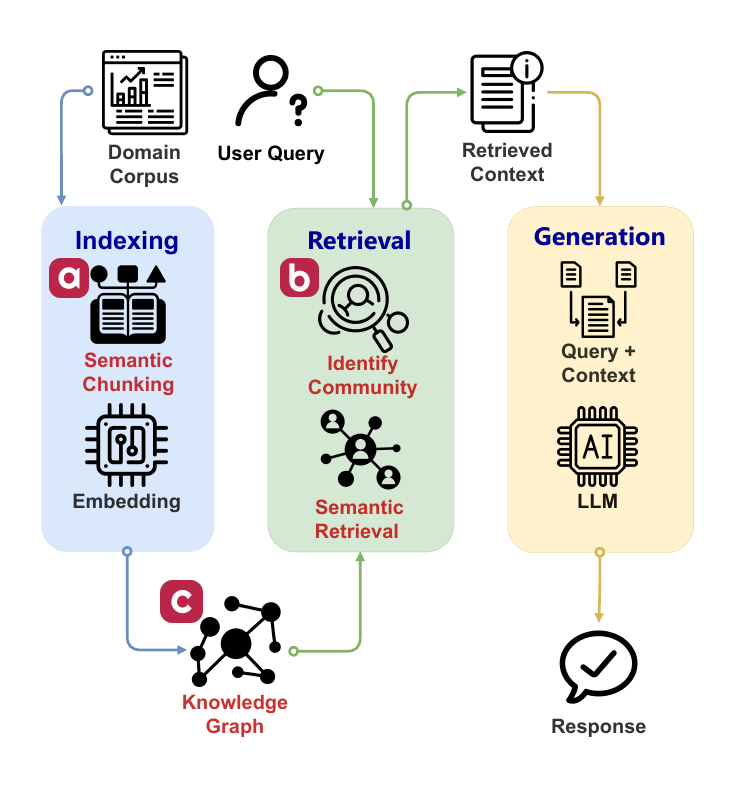}
  \centering
  \caption{SemRAG framework leveraging semantic chunking and knowledge graphs for enhanced contextual understanding; (a) Semantic Indexing: Segment the document into smaller chunks indexed and stored in a database. (b) Context Retrieval: Retrieve the top k community based on semantic similarity. (c) Knowledge Graph: Information is extracted to build the knowledge graph hierarchy with communities.}
\label{fig:improveRAG}
\end{figure}

\section{Introduction}
The recent advances of Large Language Model (LLM) have demonstrated broad utility across sectors, but general-purpose models often struggle with domain-specific accuracy due to limitations in contextual understanding and the prevalence of hallucinations \cite{10.1093/jla/laae003}. Although fine-tuning offers improvements, it is constrained by substantial data requirements and high computational costs. Retrieval Augmented Generation addresses these challenges by integrating a retrieval mechanism that supplements pre-trained models with relevant external knowledge, thereby enabling more accurate and contextually appropriate responses. Retrieval Augmented Generation (RAG) is the process of retrieving relevant content from external knowledge and incorporating it into the LLM generation process to produce accurate, relevant, and up-to-date responses \cite{10.5555/3495724.3496517}. The current RAG process requires improvement in three critical areas: indexing, retrieval, and generation \cite{gao2023retrievalaugmented}. The indexing stage involves computationally intensive pre-processing and chunking, where optimal chunk size presents a trade-off between noise reduction and context retention, thus affecting embedding quality. Inefficiencies in query matching and ranking often limit retrieval performance. Methods such as Hybrid Search and Semantic Ranking \cite{NEURIPS2024_db93ccb6} can improve certain query types, reduce retrieval errors, and limit irrelevant results. Although re-ranking can be implemented to enhance accuracy, it adds computational overhead and risks degradation from noisy or redundant data. The generation stage struggles with integrating retrieved chunks into prompts, balancing coherence, and preventing hallucinations. Poor prompt design can misinterpret intent, while advanced techniques like few-shot prompting improve performance at the cost of complexity. Extensive context requirements further strain resources and risk suboptimal responses if the model priorities irrelevant details.

Semantic chunking has emerged as a key trend for improving contextual performance in RAG pipelines. Research, including a study by Vectara, highlights the importance of optimizing chunking size during the indexing process to enhance RAG's effectiveness \cite{qu-etal-2025-semantic}. Semantic chunking stands out for its ability to maintain context and coherence, although it requires some computational resources. In contrast, methods like NLTK \cite{bird-loper-2004-nltk} and Spacy may fall short in accurately identifying sentence boundaries as they focus more on sophisticate rule base chunking that might be syntactically correct but semantically disjoint, on the other hand rule base chunking such as recursive chunking often struggles to preserve context and lead to fragmented and less meaningful chunks. Hence striking a balance between context preservation, semantic consistency, and computational efficiency is crucial for effective chunking strategies. 

One promising approach is the integration of Knowledge Graph (KG) as it enables efficient access to structured and unstructured information. By organizing data into entities (nodes) and relationships (edges), KG provides a structured representation that enables precise, context-aware retrieval. It improves relevance through disambiguation, comprehensive summaries, and deeper topic connections, offering superior contextual understanding and intuitive navigation \cite{hogan2020knowledge}. This makes KG more effective than keyword-based systems, especially for linking entities and delivering a holistic view of data in knowledge-rich tasks.

This paper addresses the limitations of conventional RAG by introducing SemRAG \textbf{Figure \ref{fig:improveRAG}}, which integrates knowledge graphs and semantic chunking. SemRAG enhances LLM output relevance and contextual understanding through local/global community search and semantic indexing. The main contributions of this paper are as follows:
\begin{itemize}
    \item  We propose SemRAG, a novel RAG architecture that uniquely integrates semantic chunking with local/global community-based subgraph retrieval from knowledge graphs, preserving both sentence-level coherence and knowledge structure relevance.

    \item We introduce a unique semantic indexing strategy that aligns semantically meaningful text chunks with structured knowledge graph entities and relations. This alignment enables more accurate and context-aware retrieval by capturing both surface-level semantics and deeper conceptual relationships.

    \item We demonstrate that SemRAG significantly improves answer relevance and answer correctness on two domain-specific benchmarks, outperforming existing RAG pipelines by up to 25\% observed in certain cases. 
\end{itemize}

\section{Existing/Related Work}

\subsection{Limitations of Traditional Fine-Tuning Methods}
While tradiational fine-tuning method offer efficient ways to adapt language models for specific tasks, they can be limited when handling knowledge-intensive or dynamic domains. PEFT relies on the model's pre-existing knowledge base, which may not suffice for domains requiring constant access to updated or specialized information \cite{10.1609/aaai.v37i11.26505}. Prompt-based tuning can struggle in tasks where external, domain-specific knowledge is needed beyond what the model was pre-trained on \cite {dong-etal-2024-survey}. Database augmentation addresses some of these limitations but may still face challenges in integrating and retrieving the most relevant information efficiently \cite{ram-etal-2023-context}.

In contrast, RAG enhances a model's performance by dynamically retrieving relevant external information during both fine-tuning and inference. This allows RAG to incorporate up-to-date and domain-specific knowledge directly from external sources, making it more effective in knowledge-intensive tasks. By continuously accessing and integrating new information, RAG overcomes the limitations of static fine-tuning approaches, enabling more accurate and contextually appropriate outputs.

\subsection{Retrieval-Augmented Generation (RAG)}
Retrieval-Augmented Generation (RAG) represents a significant advancement in large language models (LLMs) by addressing the limitations of static training data and mitigating issues like hallucinations. Unlike traditional models such as GPT-3, which generate responses based solely on pre-trained knowledge, RAG dynamically retrieves external information, enhancing factual accuracy and contextual relevance. The RAG process consists of three main stages: indexing, retrieval, and generation. Indexing involves extracting, pre-processing, and embedding information into a vector database to facilitate efficient search and retrieval \cite{10.5555/3495724.3496517, 10707868}. During retrieval, a user’s query is processed using various search methods, such as Hybrid Search \cite{doan-etal-2024-hybrid} for quick responses, Semantic Ranking for nuanced queries, and Vector Search for context-rich domains \cite{hwang-etal-2024-dslr}. The retrieved chunks are then ranked, often using re-ranking techniques to ensure relevance and reduce noise \cite{qin-etal-2024-large, 10.1145/3626772.3657848}. Finally, the ranked information is incorporated into the prompt, guiding the LLM to generate well-informed responses. Prompt engineering techniques like zero-shot, few-shot, and chain-of-thought prompting further refine the generated output, minimizing hallucinations and improving coherence \cite{hsia2025ragged}.

\subsection{Limitation of Retrieval-Augmented Generation (RAG)}
Despite its potential, optimizing RAG presents several challenges across indexing, retrieval, and generation. Indexing requires breaking documents into retrievable chunks, where aligning chunk sizes with user queries enhances specificity and accuracy \cite{zhong-etal-2025-mix}. Semantic chunking has emerged as a key trend for improving contextual performance in RAG pipelines \cite{10.1007/978-981-96-4288-5_17}. Research, including a study by Vectara, highlights the importance of optimizing chunking size during the indexing process to enhance RAG's effectiveness \cite{qu-etal-2025-semantic}. Semantic chunking stands out for its ability to maintain context and coherence, although it requires greater computational resources. In contrast, methods like NLTK and Spacy may fall short in accurately identifying sentence boundaries, while recursive chunking often struggles to preserve context. Striking a balance between context preservation, semantic consistency, and computational efficiency is crucial for effective chunking strategies. 
One promising approach is the integration of Knowledge Graph (KG) as it enables efficient access to structured and unstructured information. By organizing data into entities (nodes) and relationships (edges), KG provides a structured representation that enables precise, context-aware retrieval. It improves relevance through disambiguation, comprehensive summaries, and deeper topic connections, offering superior contextual understanding and intuitive navigation \cite{hogan2020knowledge}. This makes KG more effective than keyword-based systems, especially for linking entities and delivering a holistic view of data in knowledge-rich tasks.

\section{Methodology}

\subsection{Overview of Graph Retrieval-Augmented Generation}
GraphRAG enhances traditional RAG by integrating structured knowledge through knowledge graphs, modifying indexing, retrieval, and generation into graph construction, graph-guided retrieval, and graph-augmented response generation \cite{10771030}. In the indexing phase, GraphRAG builds a graph database from sources,  public knowledge graphs or proprietary data, mapping nodes, edges, and relationships to facilitate structured retrieval. The retrieval process identifies relevant graph elements—entities, paths, or subgraphs—by aligning user queries with graph semantics, ensuring precise information extraction. During generation, retrieved graph elements augment prompts, enabling the model to generate more contextually accurate and informed responses \cite{zhu-etal-2025-knowledge}. Microsoft’s advancements in GraphRAG introduce techniques such as iterative entity extraction and building hierarchical community detection via the Leiden algorithm, improving response retrieval efficiency and accuracy \cite{edge2024from}. By leveraging structured knowledge, GraphRAG enhances retrieval efficiency and the contextual depth of generated responses compared to traditional RAG methods.

\subsection{SemRAG} Inspired by Microsoft's Graph RAG \cite{edge2024from}, SemRAG advances the standard Retrieval-Augmented Generation (RAG) framework by integrating semantic indexing with knowledge graphs, thereby structuring data into semantically coherent and contextually rich chunks. As a result, it improves indexing and retrieval efficiency and has been shown to reduce hallucination in domain-specific applications such as finance \cite{barry-etal-2025-graphrag}. At the core of this approach lies semantic chunking, a process that groups sentences based on semantic similarity while leveraging the hierarchical architecture of knowledge graphs to support both localized and global information retrieval. This method is designed to optimize computational performance without sacrificing contextual accuracy. By incorporating specialized chunking and summarization techniques within the Graph-RAG pipeline, SemRAG achieves significant reductions in processing time and resource consumption. This optimized mechanism effectively lowers the computational burden typically associated with knowledge graph-based operations, increasing the scalability and practical viability of language models in large-scale, complex data environments.

\subsubsection{Semantic Chunking}
The Semantic chunking implementation focuses on incorporating a semantic chunking methodology into a lightweight Graph RAG framework, due to its adaptability and reduced computational requirements. Efforts to integrate this approach into a more resource-intensive platform proved challenging, prompting a shift to a simpler system better suited for customization. By applying cosine similarity for chunking with buffer size, a parameter that determines the number of adjacent sentences combined around a central sentence to preserve contextual coherence—this process ensures semantic integrity within chunks. Additionally, by relying on locally hosted models, the process efficiently handles domain-specific texts while minimizing overhead. By integrating the robust chunking process, streamlined embeddings, and consistent evaluations, this setup achieves a balance between accuracy, scalability, and efficient computational power.

\subsubsection{Indexing} The semantic chunking process aims to enhance the contextual understanding and retrieval capabilities of large language models (LLMs). By dynamically grouping sentences based on semantic similarity, the method ensures cohesive and contextually rich chunks that respect token limits. 

\textbf{Cosine Distance} adjacent \( n \)-neighbor sentences while respecting token limits. A semantic group chunk \( g \) is now defined as \cite {10.1145/361219.361220}:

{\begin{equation}
\begin{aligned}
g &= \left\{ c_i \;\middle|\; (d(c_i), d(c_{i+k})) = 1 - \dfrac{d(c_i) \cdot d(c_{i+k})}{\left\| d(c_i) \right\|_2 \left\| d(c_{i+k}) \right\|_2} < \tau, \; \forall k \in [0, n] \right\},
\end{aligned}
\end{equation}

where \( c_i \) and \( c_{i+k} \) are sentences or sentence groups within \( n \) neighborhood, \( d(c) \) represents the embedding of a sentence or group of sentences, and \( \tau \) is the threshold for cosine distance to preserve semantic cohesion. If a chunk \( g \) exceeds the token limit of 1024, it is split into smaller, overlapping sub-chunks that should be within 128 tokens, this can effectively group semantic similar neighboring sentences into the same chunk:
{
\begin{equation}
\begin{aligned}
g &= \bigcup_{j=1}^{m} g_j, \quad \text{where } g_j \cap g_{j+1} \neq \emptyset, \; |g_j| \leq 1024, \; \text{and } |g_j \cap g_{j+1}| = 128,
\end{aligned}
\end{equation}
}

where \( g_j \) represents the \( j \)-th sub-chunk, ensuring overlap between adjacent sub-chunks \( g_j \) and \( g_{j+1} \) for contextual continuity.

\begin{algorithm}[ht]
\caption{Semantic Chunking via LLM Embedding and Cosine Similarity}
\label{alg:semantic_chunking}
\KwInput{Document set $\mathcal{D} = \{d_1, \dots, d_N\}$; threshold $\theta$; buffer size $b$; token limit $T_{\max}$}
\KwOutput{Chunk set $\mathcal{C}$}

\ForEach{$d \in \mathcal{D}$}{
    $S \gets \text{Split}(d)$ 
    \tcp{Sentences $S = \{s_1, \dots, s_m\}$} 
    $\hat{S} \gets \text{BufferMerge}(S, b)$ \tcp{Contextual merging}
    $Z \gets \{z_i = \text{Embed}(\hat{s}_i)\}_{i=1}^{|\hat{S}|}$ 
    \tcp{LLM embeddings}
    
    \For{$i = 1$ to $|\hat{S}| - 1$}{
        $d_i \gets 1 - \cos(z_i, z_{i+1})$ \tcp{Cosine distance}
    }

    $c \gets [\ ],\ \mathcal{C}_d \gets \emptyset$\;
    \For{$i = 1$ to $|\hat{S}|$}{
        \eIf{$d_i < \theta$}{
            Append $\hat{s}_i$ to $c$\;
        }{
            $\mathcal{C}_d \gets \mathcal{C}_d \cup \{c\}$;\quad $c \gets [\ ]$\;
        }
    }

    \ForEach{$c \in \mathcal{C}_d$}{
        \If{$\text{Tokens}(c) > T_{\max}$}{
            $c \gets \text{SplitWithOverlap}(c)$
        }
    }

    $\mathcal{C} \gets \mathcal{C} \cup \mathcal{C}_d$\;
}
\Return{$\mathcal{C}$}
\end{algorithm}

\subsubsection{Retrieval}
The enhanced Graph RAG pipeline involves selecting the most contextually relevant chunks and community reports from the constructed knowledge graph to answer user queries efficiently. By leveraging local and global RAG strategies, the system identifies key entities, communities, and their relationships, then ranks and filters potential matches against the user’s prompt \cite{edge2024from}.

\textbf{Knowledge Graph Community Retrieval} is technique, employed within GraphRAG \cite{edge2024from} systems, enhances the contextual relevance of information retrieved in response to user queries. Rather than retrieving isolated facts or disconnected text passages, the system first constructs a knowledge graph that maps entities and their interrelationships, then applies community detection algorithms to identify clusters of semantically related entities and facts. When a query is issued, the system locates the most pertinent communities within the graph and extracts their summarized content. This approach ensures that the generated responses are not only contextually rich and interconnected but also semantically aligned with the user’s intent.

Given a community \( C_i = (V_i, E_i) \) in a knowledge graph, the community summary report \( S(C_i) \) is constructed as:

\begin{equation}
S(C_i) = \text{LLM}_{\text{summarize}}\left(\bigcup_{v \in V_i} s(v) \cup \bigcup_{(v_j, v_k) \in E_i} s(v_j, v_k)\right),
\end{equation}

 where \( s(v) \) represents the summary of node (Entities) \( v \in V_i \), \( s(v_j, v_k) \) represents the summary of edge (Relationships) \( (v_j, v_k) \in E_i \), and \( \text{LLM}_{\text{summarize}} \) is the large language model function used to aggregate and generate a coherent summary which later in retrieval process for better contextual searching.

\textbf{Local Graph RAG Search}: The local search method identifies entities and text chunks relevant to a user query \( Q \) and optional history \( H \), prioritizing contextually relevant information. The process is defined as:

\begin{equation}
\mathcal{D}_{\text{retrieved}}  = \text{Top}_k\left( \{ v \in \mathcal{V}, g \in \mathcal{G} \;\middle|\; \text{sim}(v, Q + H) > \tau_e \land \text{sim}(g, v) > \tau_d \} \right)
\end{equation}

where \( \mathcal{D}_{\text{retrieved}} \) contains top-ranked entities \( v \) and text chunks \( g \) from the knowledge graph \( V \) and Chunks \( G \). Relevance is determined by similarity thresholds \( \tau_e, \tau_d \), with the result constrained to fit the pre-defined window size \( L \). 

\textbf{Global Graph RAG Search}: The global search method uses the top-K most important and central community reports to generate a response, prioritizing relevant and central information. The process is defined as:

\begin{equation}
\mathcal{D}_{\text{retrieved}} = \text{Top}_k\left( \bigcup_{r \in \mathcal{R}_{\text{Top-K}}(Q)} \bigcup_{c_i \in C_r} \left( \bigcup_{p_j \in c_i} (p_j, \text{score}(p_j, Q)) \right), \text{score}(p_j, Q) \right)
\end{equation}

where \( \mathcal{D}_{\text{retrieved}} \) represents the top-K community reports selected based on their relevance to the query \( Q \) and centrality in the community hierarchy, \( C_r \) is the set of text chunks from each report \( r \). Within each chunk \( c_i \), \( p_j \) represents the points (sub-pieces), and \( \text{score}(p_j, Q) \) indicates the relevance of each point \( p_j \) to the query. The function \( \text{Top}_k \) selects the top-K most relevant points based on their scores.

\begin{figure}[ht]
  \includegraphics[width=0.8\linewidth]{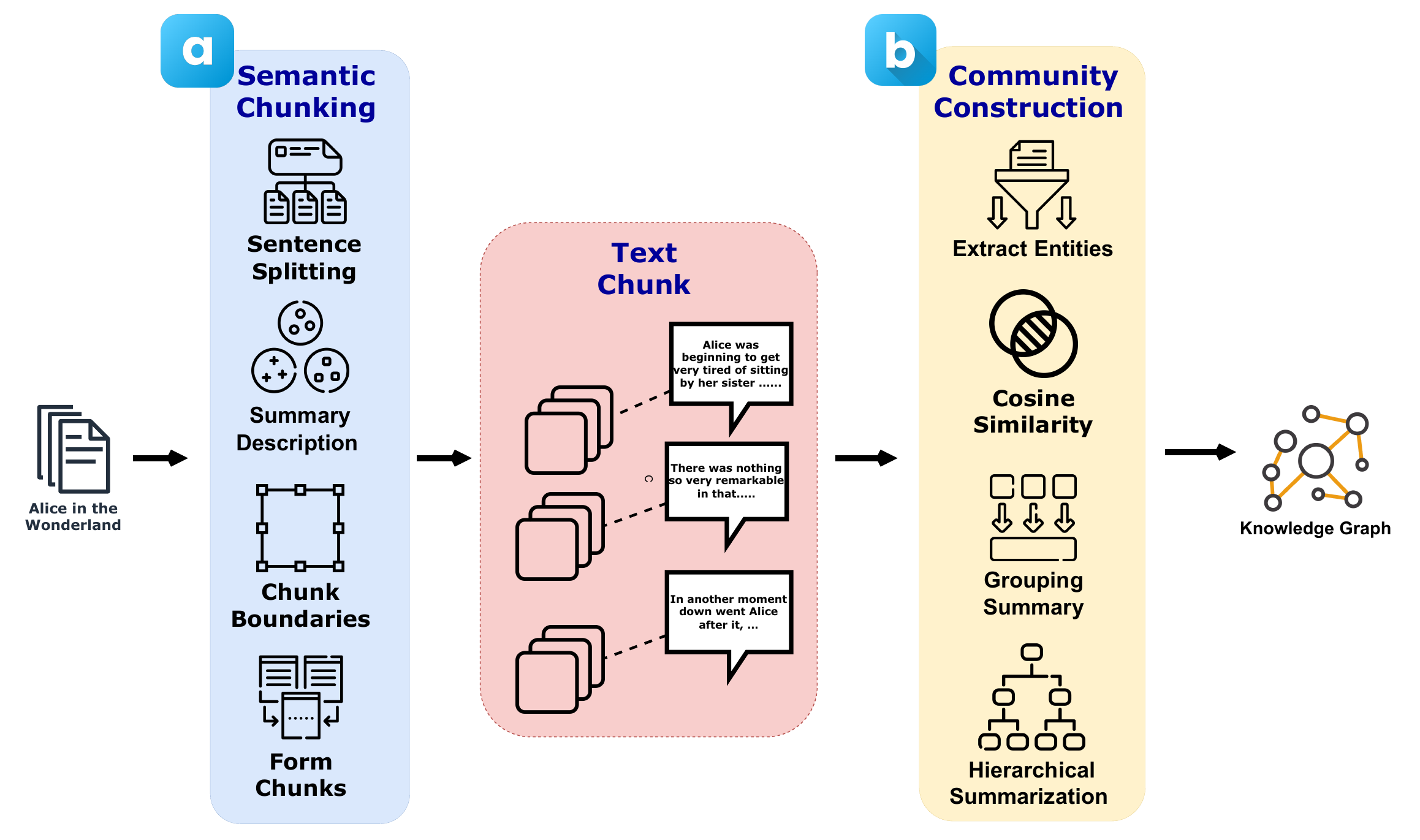}
  \centering
  \caption{SemRAG consist of two phases, semantic indexing and graph communities construction. (a) Semantic Chunking: long text has been split and segmented into meaningful chunks based on semantic relevance. (b) Graph Communities Construction: building a knowledge graph based on hierarchical community summarization.}
\label{fig: SemRAG Pipline}
\end{figure}

As shown in figure \ref{fig: SemRAG Pipline}, the semantic chunking algorithm works by first splitting the input text into individual sentences, then encoding each sentence into a vector using a pre-trained language model. It calculates the cosine similarity between each sentence and the current chunk to determine semantic closeness. If the similarity is high, the sentence is grouped with the current chunk; otherwise, a new chunk is started. This results in contextually meaningful groups of sentences. These chunks can then be used for tasks like entity extraction, summarization, and building knowledge graphs, enabling structured understanding of long, unstructured text.

\section{Experimental Setup}

\textbf{Datasets}
Two datasets are chosen to evaluate SemRAG enhancement to the Graph Rag pipeline; MultiHopRAG is a cross-domain QA dataset, which is designed to assess retrieval and reasoning across documents with associated meta data within RAG pipelines. It consists of 609 data corpus and  2,566 Q\&A, each by evidence spanning 2 to 4 documents within the data corpus. The Q\&A incorporate document metadata, representing complex, real-world scenarios often encountered in RAG applications \footnote{https://huggingface.co/datasets/yixuantt/MultiHopRAG}\cite{tang2024multihoprag}.
RAG Mini Wikipedia is a lightweight version of a Wikipedia corpus, specifically designed for the RAG pipeline. It contains 918 Q\&A pairs along with a large data corpus that is scattered across 3,200 entries. As a necessary pre-processing step, entries with similar content are merged before constructing the knowledge graph\footnote{https://huggingface.co/datasets/rag-datasets/rag-mini-wikipedia}.

\begin{table}[ht]
\centering
\caption{Comparison of Chunking Metrics Across Buffer Sizes in MultiHop and Wiki Datasets}
\vspace{4pt}
\begin{tabular}{|l|c|c|}
\hline
\textbf{Metric} & \textbf{MultiHop} & \textbf{Wiki} \\ \hline
\multicolumn{3}{|c|}{\textbf{Buffer 0}} \\ \hline
Chunks & 324 & 645 \\
Nodes  & 861 & 1728 \\
Edges  & 338 & 888 \\
Time (Sec.) & 3613 & 8249.63 \\ \hline

\multicolumn{3}{|c|}{\textbf{Buffer 5}} \\ \hline
Chunks & 292 & 765 \\
Nodes  & 600 & 2266 \\
Edges  & 65  & 1150 \\
Time (Sec.) & 5270 & 9874.03 \\ \hline

\multicolumn{3}{|c|}{\textbf{Fixed Size}} \\ \hline
Chunks & 219 & 540 \\
Nodes  & 374 & 1450 \\
Edges  & 98  & 720 \\
Time (Sec.) & 1901 & 6199.95 \\ \hline
\end{tabular}
\label{tab:vertical_buffer_comparison}
\end{table}

\subsection{Evaluation}
\textbf{Models Selection}
To evaluate the performance of SemRAG pipeline, several models are adapted to evaluate the performance with the above data sets with Mistral (7B-Instruct-Q4\_0)\cite{jiang2023mistral}, Llama3 (8B-Instruct-Q4\_0)   \cite{llama2024herd}, Gemma2(9B-Instruct-Q3\_K\_M)\cite{gemma2024improving}. The first experiment analysis the performance of each model from fixed size chunking to different semantic chunking size within the SemRAG pipeline and a deep comparison of the different models' performances with the SemRAG pipeline with semantic chunking size against baseline Naive RAG and baseline Graph RAG. In the second experiment, we further analysed the semantic chunking size impact on the LLMs' performance in different data corpus.  

\textbf{Evaluation Metrics.} Each models evaluated with Nomic-Embed-Text from Ollama for text embedding in knowledge construction and the baseline RAG vector database for compression \cite{nussbaum2024nomic}; llama3.2 for knowledge communities retrieval; ChatGPT-4–o Mini for evaluation metric to assess answer similarity, relevance, and correctness by comparing RAG-generated responses to ground-truth answers.

Retrieval-Augmented Generation Assessment (RAGAS) is used to compare baseline RAG to SemiRAG, it is a metric framework to evaluate RAG-based models, focusing on Answer Correctness, Similarity, and Relevance to ensure factual accuracy and contextual coherence \cite{es-etal-2024-ragas}. Answer Similarity computes cosine similarity between generated and ground-truth answers, while Answer Correctness balances factual and semantic overlap using an F1 score. Answer Relevance measures alignment by reverse-engineering the question from the generated response.

\begin{figure}[ht]
    \centering
    \begin{subfigure}{0.49\textwidth}
        \centering
        \includegraphics[width=\textwidth]{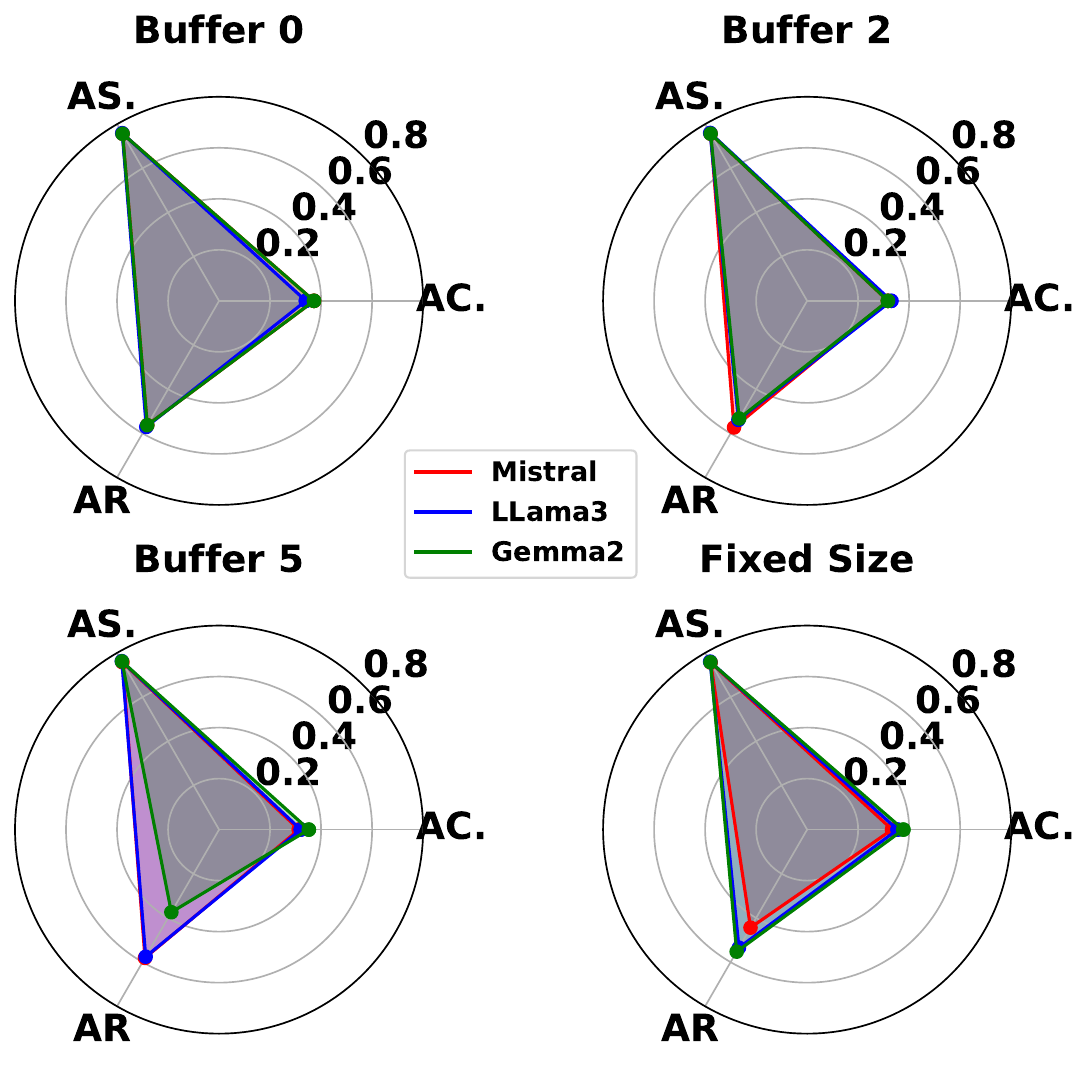}
        \captionsetup{justification=centering}
        \caption{Different Large Language Models' Performance with different Buffer Size}
        \label{fig:LLMs_Performance_and_Buffer}
    \end{subfigure}
    \hfill
    \begin{subfigure}{0.49\textwidth}
        \centering
        \includegraphics[width=\textwidth]{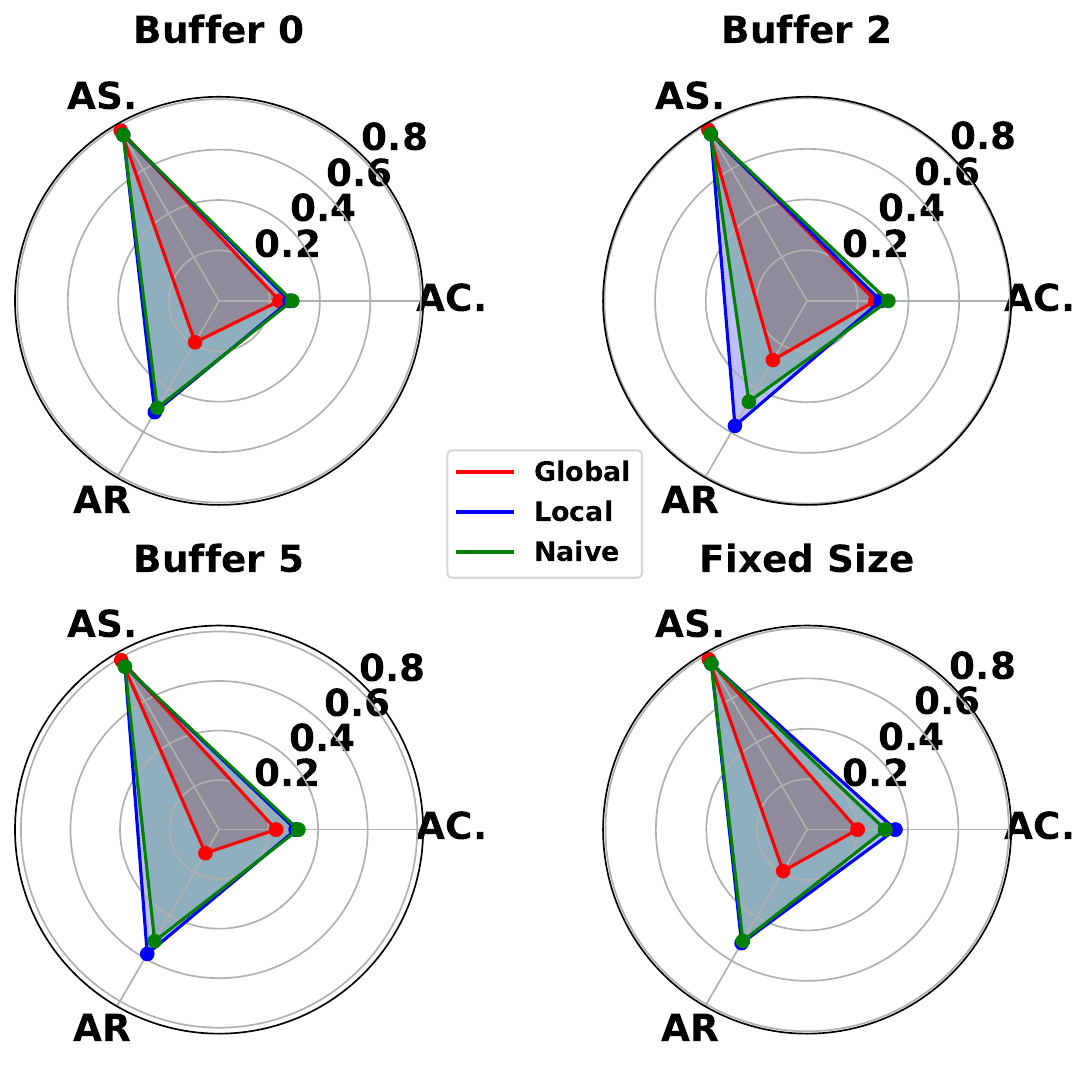}
        \captionsetup{justification=centering}
        \caption{Mistral Performance with different Buffer Size and Retrieval Methods}
        \label{fig:Mistral_Performance_and_Buffer}
    \end{subfigure}
    \captionsetup{justification=centering}
    \caption{Comparison of LLMs and Mistral Performance with Different Buffer Sizes \small (AS = Answer Similarity, AR = Answer Relevancy, AC = Answer Correctness)}
    \label{fig:LLMs_vs_Mistral_Buffer_Performance}
\end{figure}

As shown in Figure \ref{fig:LLMs_Performance_and_Buffer}, adding context significantly improves LLM performance, with the biggest jump from Buffer 0 to Buffer 2. LLaMA3 and Gemma2 outperform Mistral, with LLaMA3 excelling in correctness, as LLaMA3 with a large buffer is the best configuration, while Mistral with Global retrieval and a sufficient buffer is the most efficient alternative. Detail improvement can be seen in Figure \ref{fig:Mistral_Performance_and_Buffer}, as Mistral benefits significantly from retrieval strategies, with Global retrieval yielding the best results, while Naïve retrieval performs the worst.

\subsection{Model Performance with SemRAG}

As shown in Figure \ref{fig:Semantic_vs_naive}, Both Gemma2 and Llama3 exhibit higher scores in answer correctness with the Semantic Chunking Size 0 compared to the Naive RAG approach. Furthermore, the semantic buffer size 0 has shown significant improvement in answer relevancy across all models, which indicates a robust semantic chunking size leading to a more cohesive, informative chunk for LLMs to interpret, resulting in increasing answer relevancy performances. 

\begin{figure}[ht]
    \centering
    \begin{subfigure}[b]{0.5\textwidth}
        \centering
        \includegraphics[width=\linewidth]{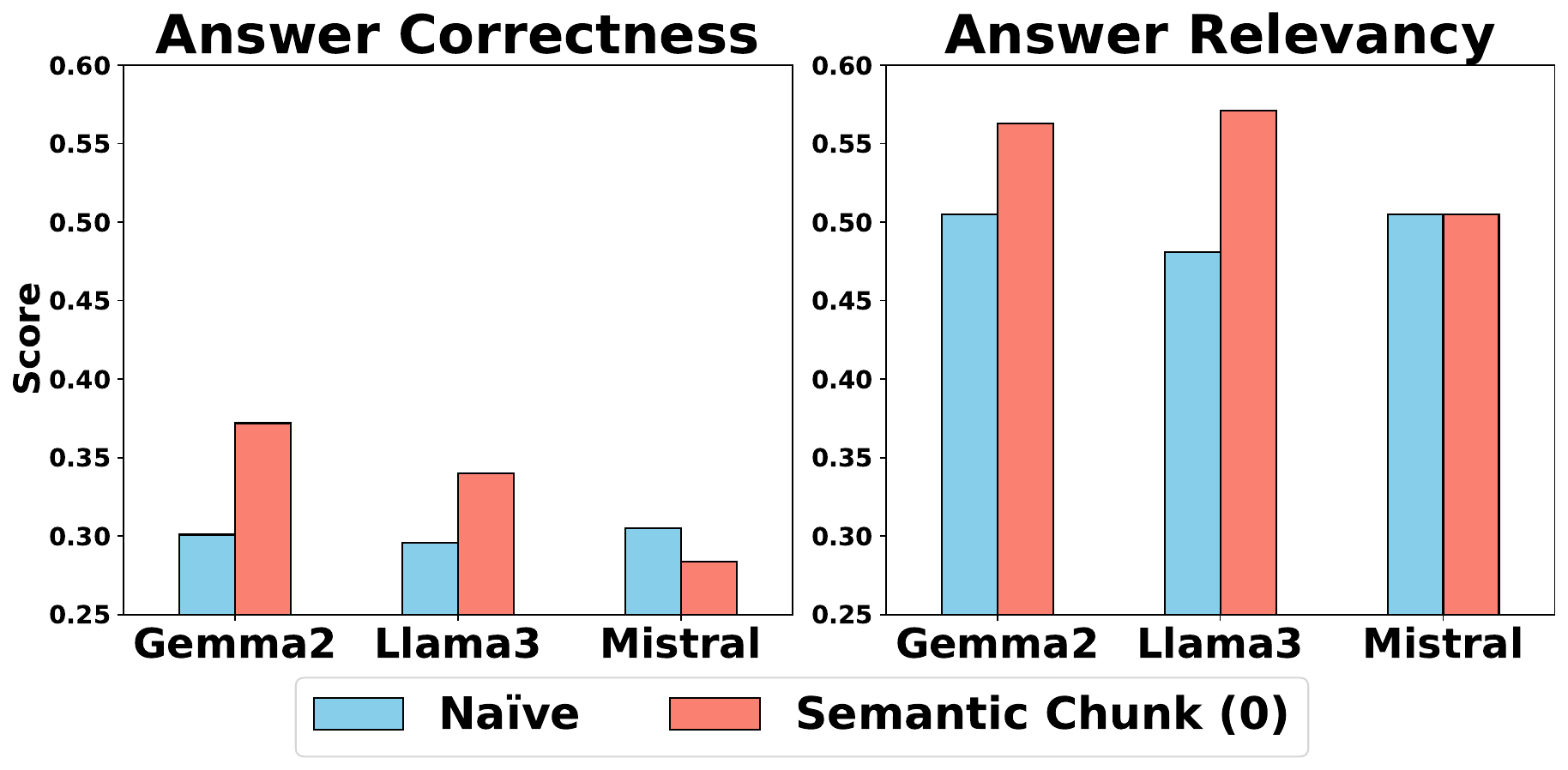}
        \caption{Semantic Chunk 0 vs Naive}
    \end{subfigure}%
    \begin{subfigure}[b]{0.5\textwidth}
        \centering
        \includegraphics[width=\linewidth]{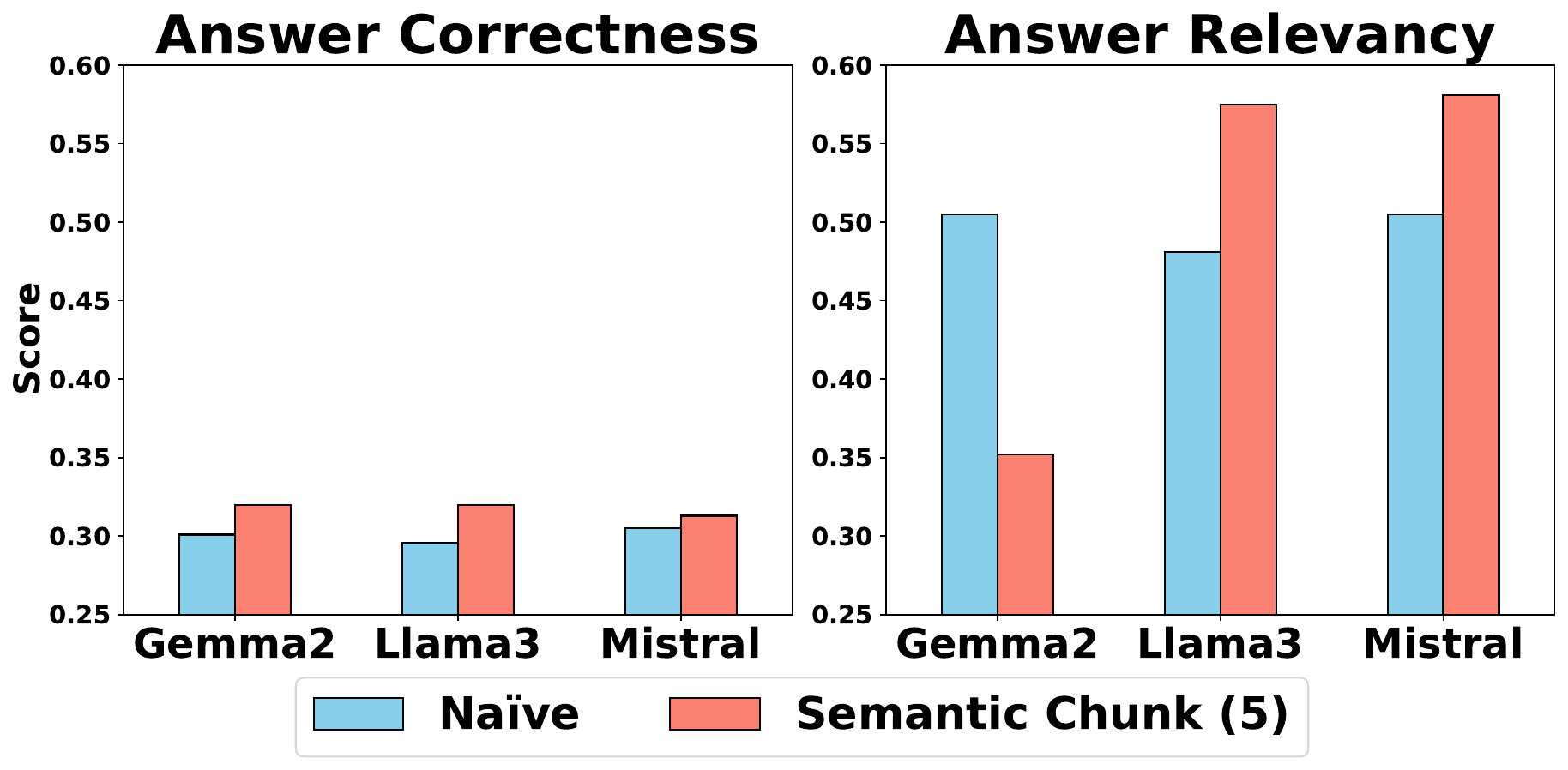}
        \caption{Semantic Chunk 5 vs Naive}
    \end{subfigure}
    \captionsetup{justification=centering}
    \caption{RAGAS metrics comparison with Multi Hop-RAG (Naive RAG vs Semantic Chunking)}
    \label{fig:Semantic_vs_naive}
\end{figure}

Based on Figure \ref{fig:Semantic_vs_fixed}, semantic Chunking with buffer size 0 generally achieves higher Answer Relevancy scores than Fixed-Size Chunking, especially for models like Llama3 and Mistral. However, Fixed-Size Chunking consistently outperforms Semantic Chunking in Answer Correctness for all models. At buffer size 5, Semantic Chunking exhibits a similar trend, with improved relevancy but lagging correctness compared to Fixed-Size Chunking.

\begin{figure}[ht]
    \centering
    \begin{subfigure}[b]{0.49\textwidth}
        \centering
        \includegraphics[width=\linewidth]{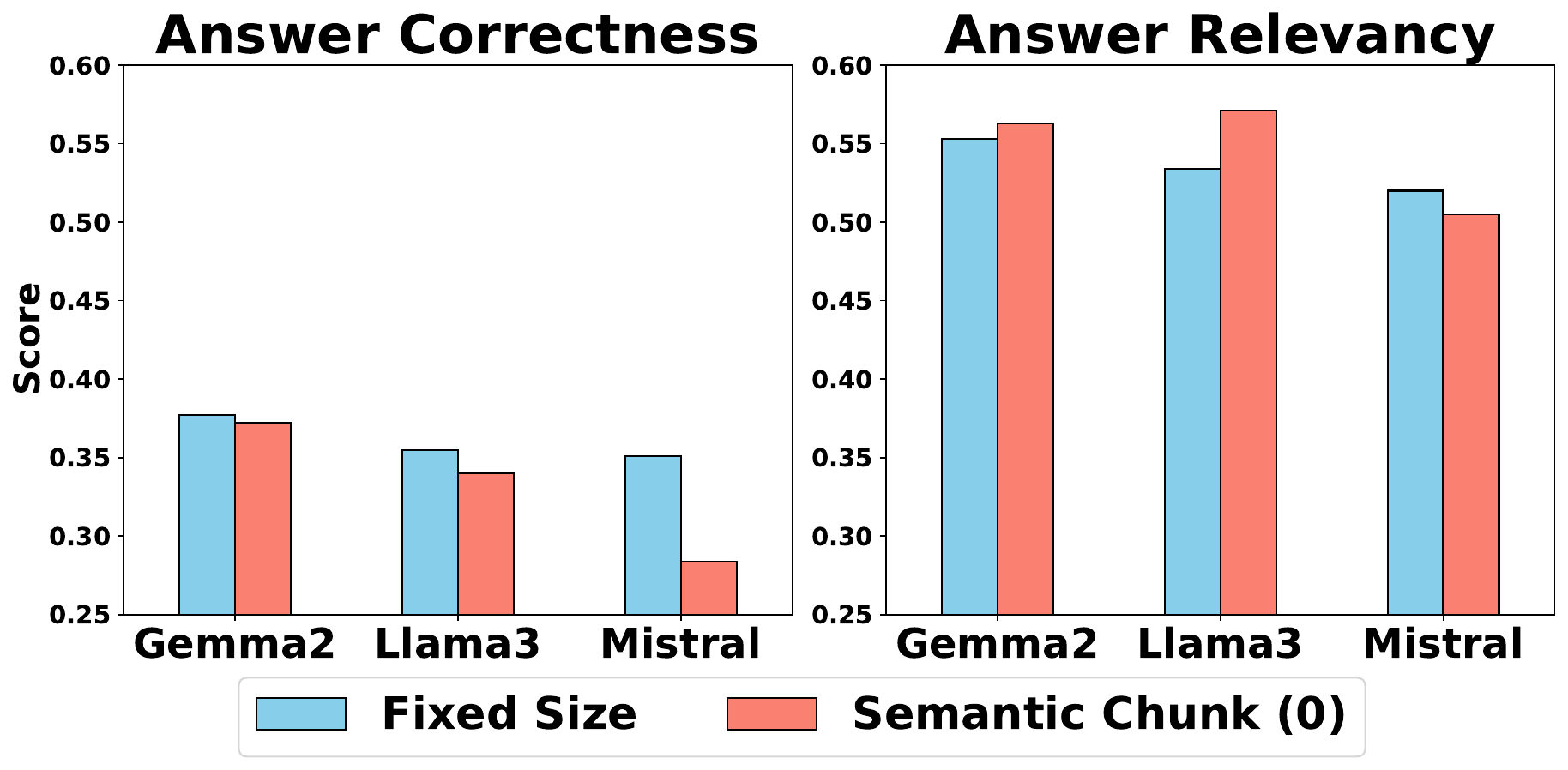}
        \caption{Semantic Chunk 0 vs Fixed Size}
    \end{subfigure}
    \hfill
    \begin{subfigure}[b]{0.49\textwidth}
        \centering
        \includegraphics[width=\linewidth]{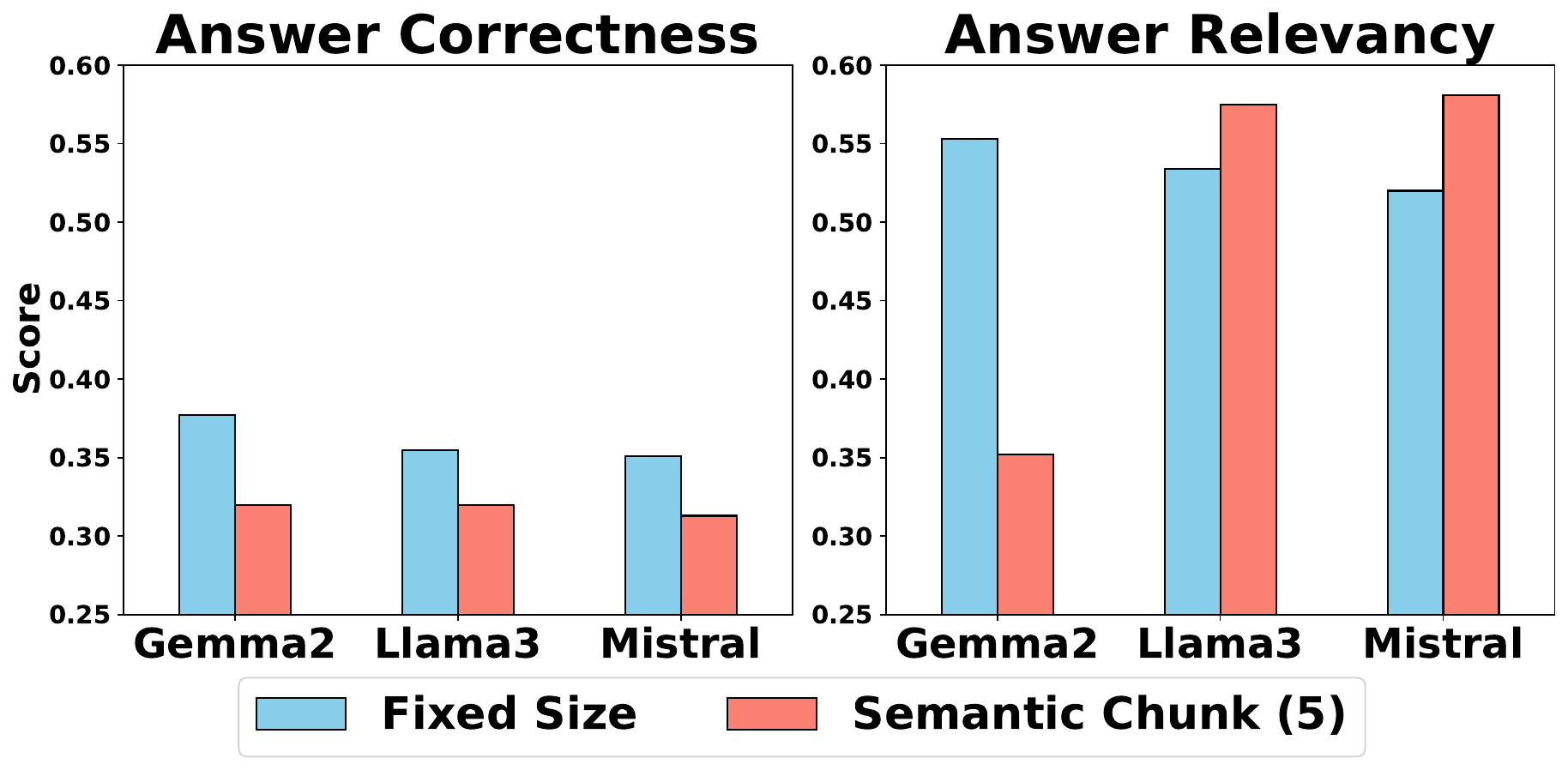}
        \caption{Semantic Chunk 5 vs Fixed Size}
    \end{subfigure}
    \captionsetup{justification=centering}
    \caption{RAGAS metrics comparison with Multi Hop-RAG (Fixed Size vs Semantic Chunking with Knowledge Graph Integration}
    \label{fig:Semantic_vs_fixed}
\end{figure}

Furthermore, the performance of Mistral in SemRAG highlights their substantial capabilities, with Local Method exhibiting great performance strengths across varying buffer sizes and configurations in figure \ref{fig:Mistral_Performance_and_Buffer}, showing the benefits of semantic chunking, with Fixed Size is best for correctness, while Semantic Chunk (5) boosts relevancy but slightly lowers correctness as more relevance information is being retrieved. However, specific configurations revealed significant limitations in semantic chunking, as some values for Llama 3 and Gemma dropped to zero with small sematic buffer size. This indicates that the models often failed to answer questions when smaller chunk sizes were used, due to insufficient information retrieval by the RAG. As a result, these models frequently generated "Insufficient Information" outputs, particularly in MultiHopRAG tasks that require the integration of multiple pieces of information. Finally, the data reveal the critical role of optimizing retrieval settings within RAG to ensure a balance between information sufficiency and specificity, ultimately supporting language models in generating accurate and reliable answers.

\subsection{Semantic Chunking and Buffer Size}

\begin{figure}[ht]
  \centering
  \hspace*{-0.1\textwidth}
  \includegraphics[width=0.9\textwidth]{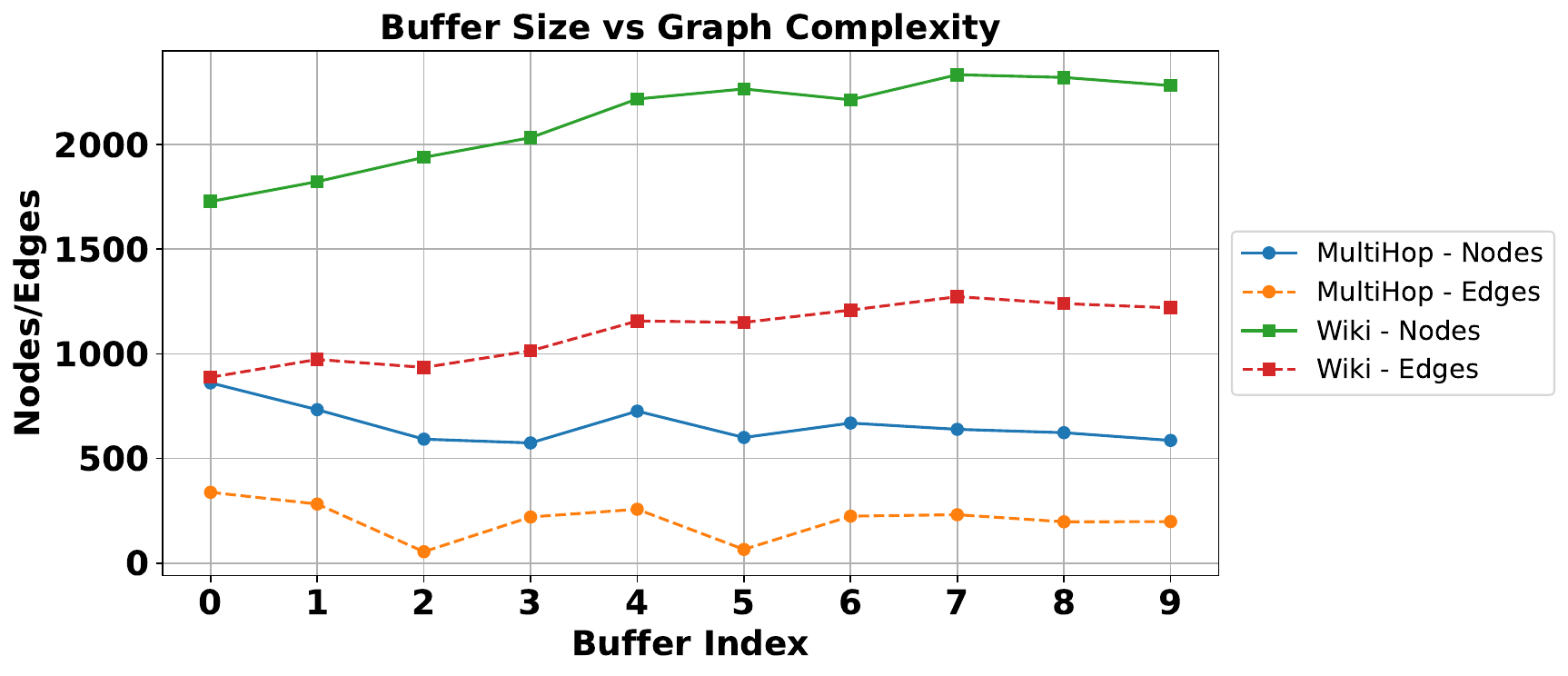}
  \captionsetup{justification=centering}
  \caption{Buffer Size vs Graph Complexity}
   \captionsetup{justification=centering,margin=2cm}
  \label{fig: Buffer Size vs Graph Complexity}
\end{figure}

\noindent\textbullet\ \textbf{Buffer size has a direct and measurable impact on graph complexity} As buffer size increases, both the number of nodes and edges in the resulting knowledge graphs scale linearly, reflecting the expanded scope of retrieved context. This effect is especially pronounced in multi-hop datasets, where even modest buffer sizes produce denser and more interconnected graphs compared to simpler Wiki datasets. The increased complexity in multi-hop tasks suggests that relevant context is more distributed and relational, making graph structure crucial for capturing meaningful connections. However, larger graph complexity does not always translate to better performance—beyond a certain buffer threshold, gains in answer accuracy plateau or decline, indicating the need for carefully optimized buffer sizes to balance retrieval depth and computational efficiency.

\begin{figure}[ht]
  \centering
  \includegraphics[width=0.8\linewidth]{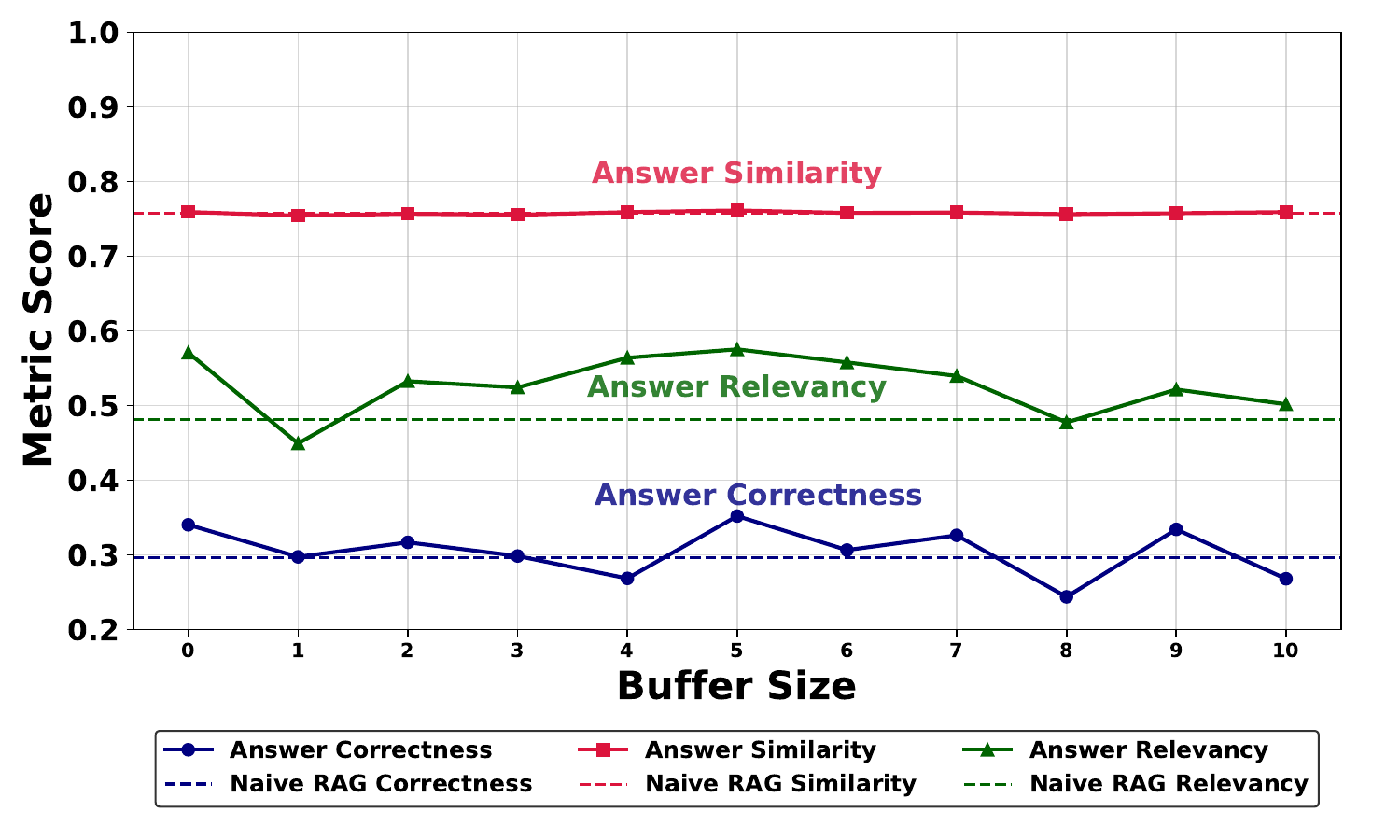}
  \captionsetup{justification=centering}
  \caption{Performance of Buffer Sizes (0-10) vs Naive RAG Based on RAGAS Test Metrics Llama 3 (Multi-Hop)}
   \captionsetup{justification=centering,margin=2cm}
  \label{fig: Performance of Buffer Sizes (0-10) vs Naive RAG Based on RAGAS Test Metrics Llama 3 (Multi-Hop)}
\end{figure}

Figure \ref{fig: Performance of Buffer Sizes (0-10) vs Naive RAG Based on RAGAS Test Metrics Llama 3 (Multi-Hop)} shows semantic chunking performance improves with buffer size, peaking at size 5 for both Answer Relevance and Correctness. This aligns with the dataset’s structure—news articles with sub-100-word sentences—allowing optimal context preservation and minimal noise. At buffer size 5, semantic chunking outperforms the naive baseline by better capturing natural content segmentation and improving knowledge graph coherence. Future buffer tuning may be required to adapt to varying dataset profiles.

\begin{figure}[ht]
  \centering
  \includegraphics[width=0.9\linewidth]{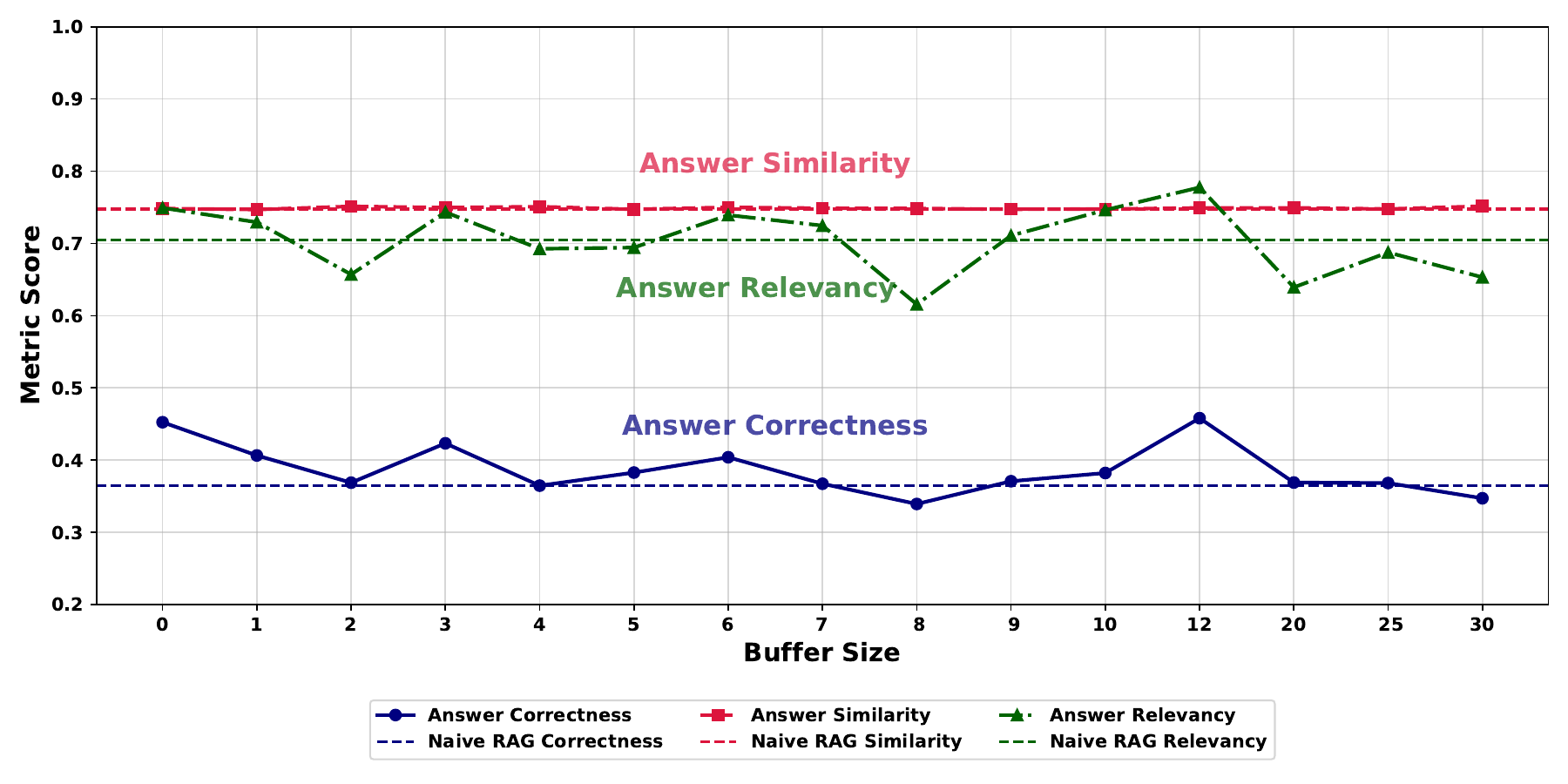}
  \captionsetup{justification=centering}
  \caption{Performance of Buffer Sizes (0-30) vs Naive RAG Based on RAGAS Test Metrics Llama 3 (wiki)}
   \captionsetup{justification=centering,margin=2cm}
  \label{fig: Performance of Buffer Sizes (0-30) vs Naive RAG Based on RAGAS Test Metrics Llama 3 (wiki)}
\end{figure}

As shown in Figure \ref{fig: Performance of Buffer Sizes (0-30) vs Naive RAG Based on RAGAS Test Metrics Llama 3 (wiki)}, the Wiki Dataset, Answer Correctness peaks at buffer size 12 before gradually declining, indicating that excessively large buffers may reduce performance. Answer Similarity remains stable (~0.8) across buffer sizes for both semantic chunking and Naive RAG, suggesting minimal sensitivity to buffer variation. Answer Relevancy shows moderate fluctuations but stays near 0.7 for semantic chunking, with Naive RAG slightly lower. These results highlight buffer size as a key factor for correctness and relevancy, with limited impact on similarity

\begin{figure}[ht]
\centering
  \includegraphics[width=0.8\linewidth]{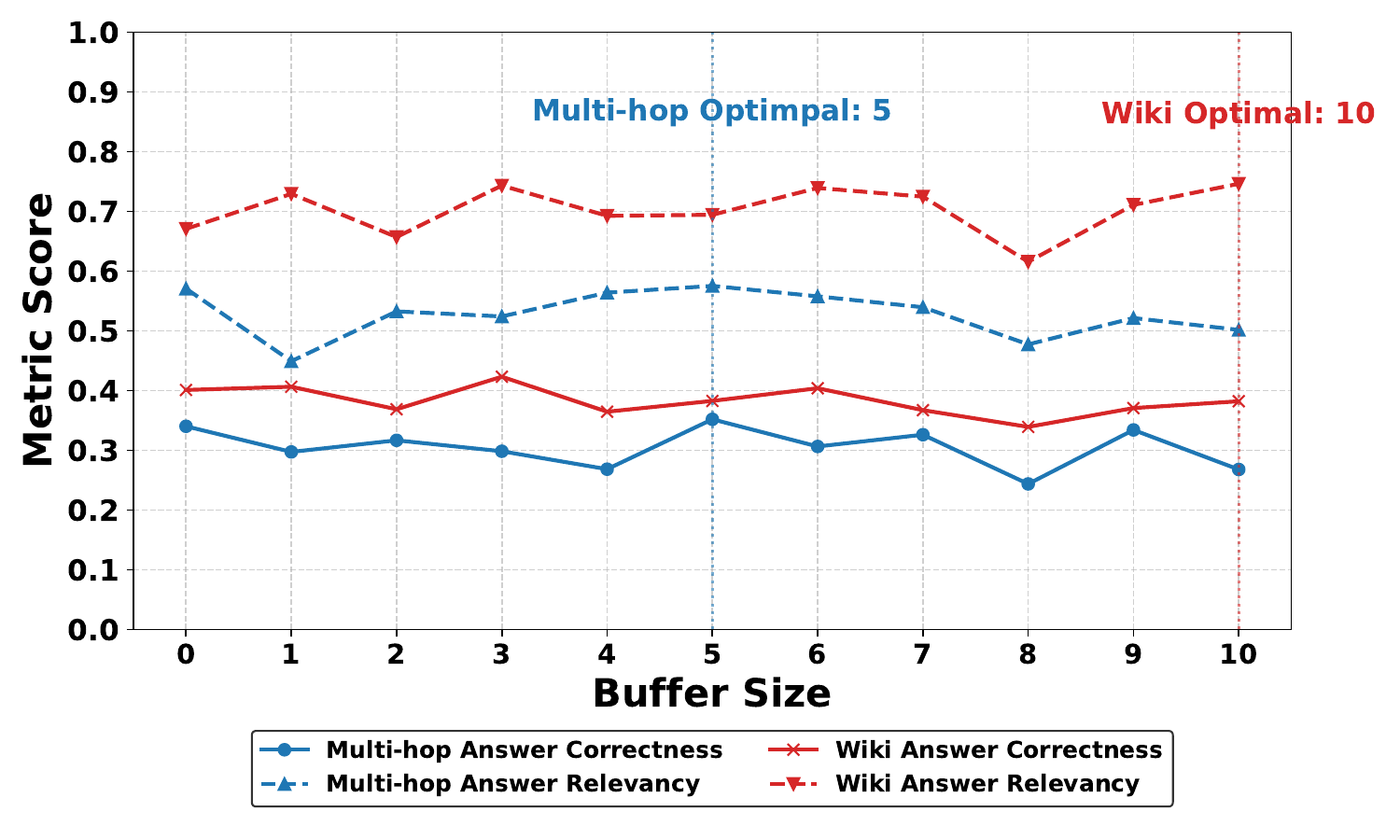}
  \caption{Optimal Chunk Comparison (Mixtral) (Multi-hop vs Wiki Dataset)}
\label{fig:Optimal Chunk Comparison (Multi-hop vs Wiki Dataset)}
\end{figure}

\noindent\textbullet\ \textbf{Buffer Size Optimization in Retrieval-Augmented Generation: Non-Linearity, Trade-offs, and Corpus Sensitivity} Figure \ref{fig:Optimal Chunk Comparison (Multi-hop vs Wiki Dataset)} highlights a non-linear relationship between buffer size and RAG performance, particularly in the metrics of answer relevancy and correctness. Contrary to the assumption that increasing buffer size uniformly enhances model outputs, results reveal dataset-specific optima: the Multi-hop dataset reaches peak performance at buffer size 5, while the Wiki dataset achieves its highest scores at size 10. This disparity underscores the need for corpus-sensitive buffer tuning, as larger buffers may introduce extraneous or tangential information that dilutes the precision of generated answers.

Across both datasets, buffer size demonstrates a positive correlation with answer quality (correctness, relevancy, and similarity) up to a critical threshold. In the initial range—buffer sizes 0–5 for Multi-hop and 0–10 for Wiki—models benefit from enriched contextual information that supports improved reasoning and content generation. However, beyond these optimal points, performance either plateaus or declines. This is likely due to information overload, where the generative model is burdened by excessive or irrelevant context, diminishing its ability to prioritize salient facts. Thus, optimal buffer sizing must be approached not as a maximization task, but as a balance between context richness and information sparsity.

Dataset structure plays a central role in determining optimal buffer configurations. In the Wiki corpus, where documents are often lengthy and context is distributed across many sentences, larger buffers are beneficial in providing necessary breadth. Conversely, the Multi-hop dataset features shorter, semantically denser passages, where smaller buffers help maintain focus and reduce the inclusion of irrelevant or redundant information. This distinction reinforces that maximizing buffer size is not universally effective; instead, it must be carefully aligned with the semantic and structural characteristics of the source corpus.

The table \ref{table:ragas_time_Wiki} of RAGAS test metrics and time for knowledge graph construction within SemRAG across different buffer sizes reveals a clear trade-off between accuracy and efficiency. Larger buffers improve answer correctness and relevancy by providing more context, but they also significantly increase KGs constitutions time. The data shows that moderate buffer sizes (around 4–7) achieve the best balance, yielding the highest correctness (\char`~0.326) (buffer 7) and strong relevancy (\char`~0.575) (Buffer 5) without excessive delays. In contrast, very large buffers lead to diminishing returns, with correctness slightly dropping due to potential context overload, while response time becomes impractically long. Therefore, choosing an optimal buffer size is crucial—too small sacrifices accuracy, while too large slows performance. A mid-range buffer provides the best trade-off, ensuring high-quality responses with reasonable speed.

Hence, the above findings highlight that a one-size-fits-all approach to semantic buffer size is insufficient for optimizing GraphRAG performance. Buffer sizes must be carefully calibrated to the characteristics of the data corpus to avoid excessive content that may introduce noise and negatively impact both relevancy and correctness. The results indicate that the optimal chunk size varies based on the corpus’s semantic density and structure. For dense, cohesive datasets, larger chunks can enhance performance by capturing interconnected information. In contrast, smaller chunks are more effective for less cohesive datasets, such as news articles, as they help reduce noise and less computational overhead \cite{wang2025speculative}. While semantic chunking generally outperforms fixed-size methods in terms of retrieval relevance, its effectiveness is highly dependent on the semantic properties of the corpus. This is further supported by the findings of Derya et al., whose research demonstrates that dynamic chunking strategies—segmenting text based on semantic coherence rather than fixed lengths—significantly improve contextual integrity, resulting in more accurate retrieval and higher-quality generated responses across varied datasets \cite{tanyildiz-etal-2024-dynamic}. Enhancing chunking with methods like semantic structuring or NLP techniques \cite{bird-loper-2004-nltk}can further improve performance, especially when the data structure is not well-suited to conventional chunking. Tailoring chunking strategies to the specific needs of each corpus boosts the accuracy and relevancy of the Graph RAG pipeline across diverse datasets. 

\section{Conclusion}
This paper introduces SemRAG, a framework that enhances conventional RAG by integrating knowledge graph and semantic chunking algorithms. Semantic RAG (SemRAG) offers notable improvements over Knowledge Graph RAG (KG-RAG) by leveraging semantic chunking to enhance computational efficiency and retrieval accuracy. SemRAG segments documents into coherent chunks based on cosine similarity, preserving context while reducing redundancy, which significantly improves retrieval relevance and correctness. This approach outperforms KG-RAG in handling large datasets due to its lower computational overhead and better adaptability in resource-constrained environments. While KG-RAG excels in capturing intricate relationships between entities, SemRAG strikes a balance by achieving superior contextual understanding and answer relevancy without the complexity and scalability challenges associated with maintaining large knowledge graphs. Overall, SemRAG provides a more efficient and practical solution for domain-specific tasks. It offers a scalable and computationally efficient method for integrating domain-specific knowledge into LLMs, resulting in an 11\% to 12\% improvement in answer relevancy as illustrated in table \ref{table:performance_metrics}. Compared to conventional RAG methods, SemRAG exhibits superior performance in answer relevancy, correctness, and similarity metrics in tested LLMs, particularly when employing optimized chunking sizes, as shown in Figure \ref{fig:Semantic_vs_naive}.

\section{Future Work}

Future work should explore lightweight approaches to integrating knowledge graphs in RAG systems to reduce computational overhead without compromising answer quality. The Nano Graph RAG pipeline used in this study exemplifies a customizable, resource-efficient solution. Emerging frameworks like Light RAG further streamline this approach with hybrid search methods to enhance retrieval efficiency and accuracy \cite{hkuds2024lightrag}.

Another key direction is developing a ground-truth metric for evaluating chunk boundaries, enabling more precise and cohesive data segmentation. This would improve semantic chunking by minimizing noise and optimizing chunk size for higher answer relevance and correctness.

Argentic chunking, which isolates atomic facts for entity-aware retrieval, offers improved precision over traditional semantic chunking. Though more computationally intensive, it enables fine-grained, context-rich retrieval when supported by advanced LLMs \cite{chen-etal-2024-dense}.

Overall, adapting chunking strategies and knowledge graph integration to the semantic structure of the data can yield more efficient, accurate, and scalable RAG systems \cite{qu-etal-2025-semantic}.

\section*{Acknowledgments}
Firstly, I would like to express my deepest gratitude to Dr. Basem Suleiman, my thesis supervisor, for the invaluable guidance, support and encouragement throughout the entire thesis project. His insights and expertise were crucial to the completion of this thesis. I am also thankful to Arthur Chen from UNSW for providing the necessary resource and provide valuable feedback and insight on the bigger picture.

To the academic staff at the School of Electrical and Computer Engineering, thanks for the teaching, guidance and the patient for my past 4 years in Sydney Uni.

Finally, I am sincerely grateful to my family members and friends for their unwavering support and encouragement throughout my undergraduate studies. Without their love and understanding, this thesis would not have been possible.

\newpage
\bibliographystyle{unsrt}  
\bibliography{references}  

\newpage

\section{Appendix}
\begin{table}[ht]
\centering
\resizebox{\textwidth}{!}{%
\begin{tabular}{|c|c|ccc|ccc|ccc|ccc|}
\hline
\textbf{Model} &  & \multicolumn{3}{c|}{Buffer 0} & \multicolumn{3}{c|}{Buffer 2} & \multicolumn{3}{c|}{Buffer 5} & \multicolumn{3}{c|}{Fixed-Size} \\ \hline
 & & Global & Local & Naive & Global & Local & Naive & Global & Local & Naive & Global & Local & Naive \\ \hline

\multirow{4}{*}{\textbf{Mistral}} 
& Chunks & 324 & 324 & 324 & 282 & 282 & 282 & 292 & 292 & 292 & 219 & 219 & 219 \\
& Nodes & 170 & 170 & 170 & 157 & 157 & 157 & 114 & 114 & 114 & 73 & 73 & 73 \\
& Edges & 65 & 65 & 65 & 72 & 72 & 72 & 36 & 36 & 36 & 28 & 28 & 28 \\
& Time (Sec.) & 8166 & 8166 & 8166 & 2633 & 2633 & 2633 & 8593 & 8593 & 8593 & 1872 & 1872 & 1872 \\ \hline

\multirow{4}{*}{\textbf{Llama3}} 
& Chunks & 324 & 324 & 324 & 282 & 282 & 282 & 292 & 292 & 292 & 219 & 219 & 219 \\
& Nodes & 861 & 861 & 861 & 586 & 586 & 586 & 661 & 661 & 661 & 374 & 374 & 374 \\
& Edges & 338 & 338 & 338 & 204 & 204 & 204 & 214 & 214 & 214 & 98 & 98 & 98 \\
& Time (Sec.) & 3613 & 3613 & 3613 & 3136 & 3136 & 3136 & 3435 & 3435 & 3435 & 1902 & 1902 & 1902 \\ \hline

\multirow{4}{*}{\textbf{Gemma2}} 
& Chunks & 324 & 324 & 324 & 282 & 282 & 282 & 292 & 292 & 292 & 219 & 219 & 219 \\
& Nodes & 769 & 769 & 769 & 592 & 592 & 592 & 600 & 600 & 600 & 315 & 315 & 315 \\
& Edges & 65 & 65 & 65 & 54 & 54 & 54 & 65 & 65 & 65 & 42 & 42 & 42 \\
& Time (Sec.) & 5528 & 5528 & 5528 & 5271 & 5271 & 5271 & 5270 & 5270 & 5270 & 2362 & 2362 & 2362 \\ \hline

\end{tabular}
}
\caption{Chunks, Graph Structure, and Processing Time for Mistral, Llama3, and Gemma2}
\label{tab:clean_combined}
\end{table}

\begin{table}[ht]
\centering
\renewcommand{\arraystretch}{1.2} 
\setlength{\tabcolsep}{6pt} 

\caption{Performance Metrics Across Different Models with SemRAG (MultiHopRAG), The best score is highlighted in blue and the ($\pm$) denotes the standard deviation of the each score.}
\label{table:performance_metrics}

\resizebox{0.7\textwidth}{!}{
\begin{tabular}{l|l|c|c|c}
\toprule
\multicolumn{2}{c|}{} & \textbf{Answer Correctness} & \textbf{Answer Similarity} & \textbf{Answer Relevancy} \\ 
\midrule

\multicolumn{5}{c}{\textbf{Fixed Size}} \\ 
\midrule
\multirow{3}{*}{\textbf{}} 
& Mistral  & 0.351 (±0.015) & 0.762 (±0.008) & 0.520 (±0.012) \\ 
& Llama3   & 0.355 (±0.012) & 0.761 (±0.007) & 0.534 (±0.011) \\ 
& Gemma2   & \textcolor{blue}{\textbf{0.377 (±0.014)}} & 0.758 (±0.009) & 0.553 (±0.013) \\ 
\midrule

\multicolumn{5}{c}{\textbf{Naïve}} \\ 
\midrule
\multirow{3}{*}{\textbf{}} 
& Mistral  & 0.305 (±0.016) & 0.756 (±0.009) & 0.505 (±0.011) \\ 
& Llama3   & 0.296 (±0.014) & 0.757 (±0.008) & 0.481 (±0.012) \\ 
& Gemma2   & 0.301 (±0.015) & 0.756 (±0.009) & 0.505 (±0.011) \\ 
\midrule

\multicolumn{5}{c}{\textbf{Semantic Chunk (0)}} \\ 
\midrule
\multirow{3}{*}{\textbf{}} 
& Mistral  & 0.284 (±0.017) & 0.756 (±0.009) & 0.505 (±0.012) \\ 
& Llama3   & 0.340 (±0.013) & 0.759 (±0.008) & 0.571 (±0.011) \\ 
& Gemma2   & 0.372 (±0.014) & 0.756 (±0.009) & 0.563 (±0.012) \\ 
\midrule

\multicolumn{5}{c}{\textbf{Semantic Chunk (5)}} \\ 
\midrule
\multirow{3}{*}{\textbf{}} 
& Mistral  & 0.313 (±0.015) & 0.758 (±0.008) & \textcolor{blue}{\textbf{0.581 (±0.013)}} \\ 
& Llama3   & 0.320 (±0.014) & \textcolor{blue}{\textbf{0.763 (±0.007)}} & 0.575 (±0.012) \\ 
& Gemma2   & 0.320 (±0.015) & 0.757 (±0.008) & 0.352 (±0.014) \\ 
\bottomrule
\end{tabular}
}
\end{table}

\begin{table}[ht]
\centering
\renewcommand{\arraystretch}{1.3} 
\setlength{\tabcolsep}{10pt} 
\caption{Performance Metrics with different buffer size (Wiki Data). The best score is highlighted in blue and the longest time to construct a knowledge graph is denoted in red.}
\label{table:ragas_time_Wiki}
\resizebox{0.9\textwidth}{!}{ 
\begin{tabular}{|l|c|c|c|c|}
\hline
\rowcolor{gray!20} \textbf{Buffer Size} & \textbf{Time (s) ↓} & \textbf{Answer Correctness ↑} & \textbf{Answer Similarity ↑} & \textbf{Answer Relevancy ↑} \\ \hline
Buffer 0 & 8249.63 & 0.401 & 0.749 & 0.671 \\ \hline
Buffer 1 & 8970.46 & 0.407 & 0.747 & 0.729 \\ \hline
Buffer 2 & 9316.69 & 0.369 & 0.751 & 0.657 \\ \hline
Buffer 3 & 9133.21 & 0.423 & 0.750 & 0.743 \\ \hline
Buffer 4 & 9479.71 & 0.365 & 0.751 & 0.693 \\ \hline
Buffer 5 & 9874.03 & 0.383 & 0.747 & 0.694 \\ \hline
Buffer 6 & 10244.40 & 0.404 & 0.750 & 0.739 \\ \hline
Buffer 7 & 10482.90 & 0.367 & 0.749 & 0.725 \\ \hline
Buffer 8 & 10417.80 & 0.339 & 0.748 & 0.616 \\ \hline
Buffer 9 & 10060.70 & 0.371 & 0.747 & 0.711 \\ \hline
Buffer 10 & 10637.40 & 0.382 & 0.748 & 0.747 \\ \hline
\rowcolor{gray!10} ... & ... & ... & ... & ... \\ \hline
Buffer 12 & 11319.60 & \textcolor{blue}{\textbf{0.458}} & \textcolor{blue}{\textbf{0.752}} & \textcolor{blue}{\textbf{0.778}} \\ \hline
\rowcolor{gray!10} ... & ... & ... & ... & ... \\ \hline
Buffer 20 & 12021.00 & 0.369 & 0.749 & 0.639 \\ \hline
Buffer 25 & 14418.60 & 0.368 & 0.749 & 0.688 \\ \hline
\rowcolor{gray!10} ... & ... & ... & ... & ... \\ \hline
Buffer 30 & \textcolor{red}{\textbf{16157.70}} & 0.347 & 0.747 & 0.653 \\ \hline
\end{tabular}}
\end{table}

\begin{table}[ht]
\centering
\renewcommand{\arraystretch}{1.3} 
\setlength{\tabcolsep}{10pt} 
\caption{Performance Metrics with different buffer size (MultiHop). The best score is highlighted in blue and the longest time to construct a knowledge graph is denoted in red.}
\label{table:ragas_time_multihop}
\resizebox{0.9\textwidth}{!}{ 
\begin{tabular}{|l|c|c|c|c|}
\hline
\rowcolor{gray!20} \textbf{Buffer Size} & \textbf{Time (s) ↓} & \textbf{Answer Correctness ↑} & \textbf{Answer Similarity ↑} & \textbf{Answer Relevancy ↑} \\ \hline
Buffer 0 & 3613 & 0.340 & 0.759 & 0.571 \\ \hline
Buffer 1 & 3038 & 0.297 & 0.754 & 0.449 \\ \hline
Buffer 2 & \textcolor{red}{\textbf{5271}} & 0.317 & 0.757 & 0.533 \\ \hline
Buffer 3 & 3007 & 0.299 & 0.755 & 0.524 \\ \hline
Buffer 4 & 3303 & 0.269 & 0.759 & 0.564 \\ \hline
Buffer 5 & 5270 & \textcolor{blue}{\textbf{0.320}} & \textcolor{blue}{\textbf{0.763}} & \textcolor{blue}{\textbf{0.575}} \\ \hline
Buffer 6 & 3420 & 0.307 & 0.758 & 0.558 \\ \hline
Buffer 7 & 3370 & 0.326 & 0.759 & 0.540 \\ \hline
Buffer 8 & 3474 & 0.244 & 0.756 & 0.477 \\ \hline
Buffer 9 & 3361 & 0.334 & 0.758 & 0.522 \\ \hline
Buffer 10 & 3418 & 0.268 & 0.759 & 0.502 \\ \hline
\end{tabular}}
\end{table}

\end{document}